\documentclass{article}

\usepackage{microtype}
\usepackage{graphicx}
\usepackage{subfigure}
\usepackage{booktabs} % for professional tables
\usepackage{amsmath}
\usepackage{tabularx}
\usepackage{hyperref}

\newcommand{\bs}[1] {  \boldsymbol{#1}           }
\newcommand{\ff}[1] {  \mbox{\footnotesize{#1}}  }

\usepackage[accepted]{icml2020}

\icmltitlerunning{Self-explaining Artificial Intelligence}

\begin{document}

\twocolumn[
\icmltitle{Self-explaining AI as an alternative to interpretable AI}

% It is OKAY to include author information, even for blind
% submissions: the style file will automatically remove it for you
% unless you've provided the [accepted] option to the icml2020
% package.

% List of affiliations: The first argument should be a (short)
% identifier you will use later to specify author affiliations
% Academic affiliations should list Department, University, City, Region, Country
% Industry affiliations should list Company, City, Region, Country

% You can specify symbols, otherwise they are numbered in order.
% Ideally, you should not use this facility. Affiliations will be numbered
% in order of appearance and this is the preferred way.
%\icmlsetsymbol{equal}{*}

\begin{icmlauthorlist}
\icmlauthor{Daniel C.\ Elton}{to}
\end{icmlauthorlist}

\icmlaffiliation{to}{Imaging Biomarkers and Computer-Aided Diagnosis Laboratory, Radiology and Imaging Sciences, National Institutes of Health Clinical Center, Bethesda, MD 20892, USA}

\icmlcorrespondingauthor{Daniel Elton}{daniel.elton@nih.gov}

% You may provide any keywords that you
% find helpful for describing your paper; these are used to populate
% the "keywords" metadata in the PDF but will not be shown in the document
\icmlkeywords{Interpretability,explainability,explainable artificial intelligence,XAI,trustworthiness,robustness,self-explaining AI,deep learning}

\vskip 0.3in
]

% this must go after the closing bracket ] following \twocolumn[ ...

% This command actually creates the footnote in the first column
% listing the affiliations and the copyright notice.
% The command takes one argument, which is text to display at the start of the footnote.
% The \icmlEqualContribution command is standard text for equal contribution.
% Remove it (just {}) if you do not need this facility.

\printAffiliationsAndNotice{}  % leave blank if no need to mention equal contribution
%\printAffiliationsAndNotice{\icmlEqualContribution} % otherwise use the standard text.

\begin{abstract}
While it is often possible to approximate the input-output relations of deep neural networks with a few human-understandable rules, the discovery of the double descent phenomena suggests that such approximations do not  accurately capture the mechanism by which deep neural networks work. Double descent indicates that deep neural networks typically operate by smoothly interpolating between data points rather than by extracting a few high level rules. As a result, neural networks trained on complex real world data are inherently hard to interpret and prone to failure if asked to extrapolate. To show how we might be able to trust AI despite these problems we explore the concept of self-explaining AI, which provides both a prediction and explanation.  We also argue AIs systems should include a ``warning light'' using techniques from applicability domain analysis and anomaly detection to warn the user if a model is asked to extrapolate outside its training distribution. %A video of a talk on this paper can be found \href{https://www.youtube.com/watch?v=Py7PVdcu7WY&}{\underline{here}}. \\
\end{abstract}

%--------------------------------------------------------------------------
\section{Introduction} 
There is growing interest in developing methods to explain deep neural network function, especially in high risk areas such as medicine and driverless cars. Such explanations would be useful to ensure that deep neural networks follow known rules and when troubleshooting failures. The European Union's 2016 General Data Protection Regulation says that companies must be able to provide an explanation to consumers about decisions made by artificial intelligences~\cite{Kaminski2018}, which has helped bolster growing interest on explainable AI and methods for interpreting deep neural network function. Despite the development of numerous techniques for interpreting deep neural networks, all such techniques have flaws, many of which are not well appreciated~\cite{Rudin2019,lipton2016Mythos}. More troubling, though, is that a new understanding is emerging that deep neural networks function through the brute-force local interpolation of data points, rather than global fitting procedures~\cite{Hasson2019}. This calls into question long-held narratives that deep neural networks ``extract'' high level features and rules. It also means that neural networks have no hope of extrapolating outside their training distribution. If not properly understood, current interpretability methods can lead to over optimistic projections about the generalization ability of a neural network. 

In response to difficulties raised by explaining black box models, Rudin argues for developing better interpretable models instead, arguing that the ``interpetability-accuracy'' trade-off is a myth. While it is true that the notion of such a trade-off is not rigorously grounded, empirically in many domains the state-of-the art systems are all deep neural networks. For instance, most state-of-art AI systems for computer vision are not interpretable in the sense required by Rudin. Even highly distilled and/or compressed models which achieve good performance on ImageNet require at least 100,000 free parameters~\cite{lillicrap2019does}. Moreover, some parts of the human brain appear to ``black boxes'' which perform interpolation over massive data with billions of parameters, so the interpretability of the brain using things like psychophysics experiments has come under question~\cite{Hasson2019}. If evolution settled on a model (the brain) which contains uninterpretable components, then we expect advanced AIs to also be of that type. Interestingly, although the human brain is a ``black box'', we are able to trust each other. Part of this trust comes from our ability to ``explain'' our decision making in terms which make sense to us. Crucially, for trust to occur we must believe that a person is not being deliberately deceptive, and that their verbal explanations actually maps onto the processes used in their brain to arrive at their decisions.  

Motivated by how trust works between humans, in this work we explore the idea of self-explaining AIs. Self-explaining AIs yield two outputs - the decision and an explanation of that decision. This idea is not new, and it is something which was pursued in expert systems research in the 1980s~\cite{Swartout1983expertsystem}. More recently Kulesza et al.\ introduced a model which offers explanations and studied how such models allow for ``explainable debugging'' and iterative refinement~\cite{Kulesza2015}. However, in their work they restrict themselves to a simple interpretable model (a multinomial naive Bayes classifier). A ``self-explaining'' neural network has also been developed which describes predictions using a number of human interpretable concepts or prototypes~\cite{Alvarez-MelisNIPS2018}. In a somewhat similar vein, Chen et al.~\cite{Chen2019NIPs} have proposed a ``This looks like That'' network. Unlike previous works, in this work we explore how we might create trustworthy self-explaining AI for networks and agents of arbitrary complexity. We also seek for a more rigorous way to make sure the explanation given is actually explaining an aspect of the mechanism used for prediction.  Unlike previous works this work focuses on a general approach and attempts to address concerns about robustness and AI safety.
%, including AI safety for artificial general intelligences (AGIs). 
%\cite{Wang2018TieNet}

After defining key terms, we discuss the challenge of interpreting deep neural networks raised by recent studies on interpolation and generalization in deep neural networks. Then, we discuss how self-explaining AIs might be built. We argue that they should include at least three components - a measure of mutual information between the explanation and the decision, an uncertainty on both the explanation and decision, and a ``warning system'' which warns the user when the decision falls outside the domain of applicability of the system. We hope this work will inspire further work in this area which will ultimately lead to more trustworthy AI. 

%------------------------------------------------------------ 
\section{Interpretation, explanation, and self-explanation}
As has been discussed at length elsewhere, different practitioners understand the term ``intepretability'' in different ways, leading to a lack of clarity (for detailed reviews, see\cite{lipton2016Mythos,Ahmad2018InterpretabilityHealthcare,Murdoch2019PNAS,arya2020explanation}). The related term ``explainability'' is typically used in a synonymous fashion~\cite{Rudin2019}, although some have tried to draw a distinction between the two terms~\cite{LaLondearxiv}. Here we take explanation/explainability and interpretation/interpretability to be synonymous. Murdoch et al.\ define an \textbf{explanation} as a verbal account of neural network function which is descriptively accurate and relevant~\cite{Murdoch2019PNAS}. By ``descriptively accurate'' they mean that the interpretation reproduces a large number of the input-output mappings of the model. The explanation may or may not map onto how the model works internally. Additionally, any explanation will be an approximation, and the degree of approximation which is deemed acceptable may vary depending on application. By ``relevance'', what counts as a ``relevant explanation'' is domain specific -- it must be cast in terminology that is both understandable and relevant to users. For deep neural networks, the two desiderata of accuracy and relevance appear to be in tension - as we try to accurately explain the details of how a deep neural network interpolates, we move further from what may be considered relevant to the user.

This definition of explanation in terms of capturing input-output mappings in a human understandable way contrasts with a second meaning of the term explanation which we may call \textbf{mechanistic explanation}. Mechanistic explanations abstract faithfully (but approximately) the actual data transformations occurring in the model. To consider why mechanistic explanations can be useful, consider a deep learning model we trained recently to segment the L1 vertebra in CT scans~\cite{Elton2020SPIE}. The way a radiologist identifies the L1 vertebra is by scanning down from the top of the body and finding the last vertebra that has ribs attached to it, which is T12. L1 is directly below T12. In our experience our models for identifying L1 tend to be brittle, indicating they probably use a different approach. For instance, they may do something like ``locate the bright object in the middle of the image'' or ``locate the bright object which is just above the kidneys''. These techniques would not be as robust as the technique used by radiologists. If a self-explaining AI had a model of human anatomy and could couch its explanations with reference to standard anatomical concepts, that would go a long way towards engendering trust. 

In general, the ``Rashomon Effect'' ~\cite{Breiman2001} says that for any set of noisy data, there are a multitude of models of equivalent accuracy, but which differ significantly in their internal mechanism. This is why only mechanistic explanations are helpful for addressing concerns about robustness, generalization ability, and trustworthiness. As a real-world example of the Rashomon Effect, when detecting Alzheimer's disease in brain MRI using a CNN the visualized interpretations for models trained on different train-test folds differed significantly, even though the models were of equivalent accuracy~\cite{sutre2020visualization}. Even more troubling, the visualizations differed between different runs on the same fold, with the only difference being in the random initialization of the network~\cite{sutre2020visualization}. 

We also note that interpretations can vary between test examples.\cite{barnes2018machine} In many works only a few examples (sometimes cherry-picked) are given to ``explain'' how the model works, rather than attempting to summarize the results of the interpretability method on the entire test set. It appears that in deep neural networks it is possible the mechanism of prediction can differ between models of equivalent accuracy, even when the models all have the same architecture, due to peculiarities of the training data and initialization used. On top of this issue, it is also possible that specific details of the mechanism may vary wildly within a given model across different test cases. 

There is another type of explanation we wish to discuss which we may call \textbf{meta-level explanation}. Richard P.\ Feynman said ``What I cannot create, I do not understand''. Since we can create deep neural networks, we do understand them, in the sense of Feynman, and therefore we can explain them in terms of how we build them. More specifically, we can explain neural network function in terms of four components necessary for creating them - data, network architecture, learning rules (optimization method), and objective function~\cite{Richards2019}. The way one explains deep neural network function from data, architecture, and training is analogous to how one explains animal behaviour using the theory of evolution. The evolution of architectures by ``graduate student descent'' and the explicit addition of inductive biases mirrors the evolution of organisms. Similarly, the training of architectures mirrors classical conditioning in animals. The explanation of animal behaviour in terms of meta-level theories like evolution and classical conditioning has proven to be enormously successful and stands in contrast to attempts to seek detailed mechanistic accounts. 
 
%``The incidental emergence of such rules is not the “goal” of the network and the network does not “use” the rules to extrapolate. This mindset, in fact, resembles pre-Darwinian teleological thinking and “just-so stories” in biology.''
\begin{table*}\label{methodstable}
\begin{tabularx}{\textwidth}{XX}
\textbf{Early heatmapping} \newline
Saliency maps \cite{Erhan2009ActivationMaximization,simonyan2013deep}\newline
occlusion maps \cite{zeiler2013visualizing}  \newline
deconvolution \cite{zeiler2013visualizing} \newline
guided backprop \cite{SpringenbergDBR14}\newline
Inverting CNNs \cite{DosovitskiyB16}\newline
gradient*input \cite{shrikumar2017learning}\newline
GradCAM \cite{Selvaraju_2019}\newline
iterative mapping \cite{Bordes2018}\newline
Integrated Gradient \cite{SundararajanTY17}\newline
Meaningful Perturbations \cite{DBLP:journals/corr/FongV17}\newline
\textbf{Layerwise relevance propagation based heatmapping}\newline
layerwise relevance propagation \cite{Bach2015} \newline
Pattern LRP \cite{DBLP:conf/iclr/KindermansSAMEK18}\newline
Deep Taylor Decomposition \cite{Montavon2017}\newline
DeepLIFT \cite{shrikumar2017learning}  \newline
\textbf{Shapley value based}\newline
Shapley Additive Explanations \cite{LundbergL17} \newline
Neuron Shapley \cite{ghorbani2020neuron} \newline
L-Shapley and C-Shapley \cite{ChenSWJ19} \newline
&
\textbf{Activation maximization based}\newline
Activation maximization \cite{Erhan2009ActivationMaximization}\newline
Deep Visualization \cite{DBLP:journals/corr/YosinskiCNFL15}\newline
%DeepDream \cite{deepdream}\newline
Deep Generator Networks \cite{NguyenDYBC16}\newline
\textbf{Surrogate model based}\newline
LIME \cite{Ribeiro2016}\newline
linear classifier “probes” \cite{AlainICLR17}\newline
“distilling” a neural network \cite{frosst2017distilling}\newline
Anchors \cite{DBLP:conf/aaai/Ribeiro0G18}\newline
\textbf{Referencing specific training data} \newline
influence functions \cite{Koh2017}\newline
Fischer kernels \cite{pmlr-v89-khanna19a}\newline
Representative points \cite{DBLP:conf/nips/YehKYR18}\newline
“This looks like That” \cite{Chen2019NIPs}\newline
\textbf{Others}\newline
Network Dissection \cite{Zhou2019}\newline
Kandinsky patterns\cite{Holzinger2019}\newline
Neuron deletion \cite{DBLP:conf/iclr/MorcosBRB18}\newline
%exploring failure modes \cite{Goertzel2015AGIConf}\newline
Concept activation vectors \cite{DBLP:conf/icml/KimWGCWVS18}\newline
Interpretable filters \cite{DBLP:conf/cvpr/ZhangWZ18a}\newline
explanatory graphs \cite{DBLP:conf/aaai/ZhangCSWZ18}\newline
\end{tabularx}
\caption{The interpretation method ``zoo''.}
\end{table*}

%----------------------------------------------------------- 
\section{Why deep neural networks are generally non-interpretable}
Many methods for interpretation of deep neural networks have been developed, many of which are listed in table 1. %\cite{Bordes2018}, ``distilling'' a neural network into a simpler model~\cite{frosst2017distilling}, exploring failure modes and adversarial examples~, visualizing filters in CNNs~\cite{zeiler2013visualizing}, activation maximization based visualizations~\cite{Erhan2009ActivationMaximization}, influence functions~\cite{Koh2017}, Shapley values~\cite{Shapley}, Local Interpretable Model-agnostic Explanations (LIME)~\cite{Ribeiro2016}, linear classifier ``probes'' to interpret intermediate layers~\cite{AlainICLR17}, DeepLIFT ~\cite{shrikumar2017learning}, explanatory graphs~\cite{Zhang2018AAAI}, and layerwise relevance propagation~\cite{Bach2015}. 
Yet, all of these methods capture only particular aspects of neural network function, and the outputs of these methods are very easy to misinterpret~\cite{Rudin2019,LieMasters,yeh2019infidelity}. Often the output of interpretability methods vary largely between test cases, but only a few ``representative'' cases (often hand picked) are shown in papers. Moreover, it has been shown that popular methods such as LIME~\cite{Alvarez-MelisNIPS2018}, Shapley values~\cite{Alvarez-MelisNIPS2018}, and saliency maps~\cite{DBLP:conf/icml/NieZP18,dombrowski2019explanations,yeh2019infidelity,Adebayo2018NeurIPS,Kindermans2019} are not robust to small changes in the image such as Gaussian noise. Saliency maps have additional problems - for even if many layers in a neural network are randomized, or if labels are scrambled, they show nearly the same output.\cite{Adebayo2018NeurIPS} Recently Hase et al.\ have performed tests with a large pool of human subjects to see if different explainability techniques can actually help the user predict a neural network's behaviour.\cite{hase2020evaluating} Out of the methods they compared, LIME, Anchors, Decision Boundary, and ``This looks like That'', LIME with tabular data is the only setting where an improvement was noted.\cite{hase2020evaluating} A different approach to benchmarking explainability methods was pursued by Hooker et al.\ (2020).\cite{Hooker2020} They looked a several saliency type explanations (Integrated Gradients, Guided Backprop, etc.) and looked at the features in the image that were deemed important. They then removed some of those features and retrained the network. As a baseline, they also removed features randomly.  For the conventional saliency map techniques, the features ``important'' by the explanations were actually no more important than random features.\cite{Hooker2020} While they developed a modified saliency method which was able to identify important features, the results showed the importance of rigorously testing explainability methods before fielding them.\cite{Hooker2020}

As we discussed before, we do not expect the current push towards more interpretable models led by Rudin and others to be successful in general - deep neural networks are here to stay, and they will become even more complex and inscrutable as time goes on. Lillicrap \& Kording~\cite{lillicrap2019does} note that attempts to compress deep neural networks into a simpler interpretable models with equivalent accuracy typically fail when working with complex real world data such as images or human language. If the world is messy and complex, then neural networks trained on real world data will also be messy and complex. Leo Breiman, who equates interpretability with simplicity, has made a similar point in the context of random forest models~\cite{Breiman2001}. In many domains, the reason machine learning is applied is because of the failure of simple models or because of the computational burden of physics-based simulation. While we agree with Rudin that the interpretability-accuracy trade-off is not based on any rigorous quantitative analysis, we see much evidence to support it, and in some limiting cases (for example superintelligent AGIs which we cannot understand even in principle or brain emulations, etc) the inescapability of such a trade-off existing to some extent becomes clear. 

On top of these issues, there is a more fundamental reason to believe it will be hard to give mechanistic explanations for deep neural network function. For some years now it has been noted that deep neural networks have enormous capacity and seem to be vastly underdetermined, yet they still generalize. This was shown very starkly in 2016 when in Zhang et al.\ showed how deep neural networks can memorize random labels on ImageNet images~\cite{zhang2016understanding}. More recently it has been shown that deep neural networks operate in a regime where the bias-variance trade-off no-longer applies~\cite{Belkin2019}. As network capacity increases, test error first bottoms out and then starts to increase, but then (surprisingly) starts to decrease after a particular capacity threshold is reached. Belkin et al.\ call this the ``double descent phenomena''~\cite{Belkin2019} and it was also noted in an earlier paper by Sprigler et al~\cite{Spigler2019}, who argue the phenomena is analogous to the ``jamming transition'' found in the physics of granular materials. It was also discovered to hold theoretically for the single layer perceptron by Opper et al. (1990).\cite{Opper1990} The phenomena of ``double descent'' appears to be universal to all machine learning~\cite{Belkin2019,belkin2019models}, although its presence can be masked by common practices such as early stopping~\cite{Belkin2019,nakkiran2019deep}, which may explain why it took so long to be noticed in the context of deep learning. 

In the regime where deep neural networks operate, they not only interpolate each training data point, but do so in a ``direct'' or ``robust''  way~\cite{Hasson2019}. This means that the interpolation does not exhibit the overshoot or undershoot which is typical of overfit models, rather it is almost a piecewise interpolation. The use of interpolation implies a corollary - the inability to extrapolate. 

An illuminating example of direct fitting is given by Hasson et al.\ showing direct fitting of a parabolic function with noise.\cite{Hasson2019} The computations involved are clearly local - similar to nearest neighbors type computations. Additionally, the global trend ($y \propto x^2$) is not extracted. Because of this, there is clearly no hope for extrapolation. One the other hand, the model is flexible enough to fit any data, so if the parabolic curve was to suddenly stop and turn into something completely different, such as a sine wave, it would have no issue. These observations call into question notions that deep neural networks ``extract'' high level features that are of particular interest - such as the whiskers of a cat. In actuality, it seems they are interpolating between a very large number of features, some of which are particular to the training data. Recent work confirms this idea, showing that neural networks rely heavily on ``non-robust'' features, and that this fact is actually a key to how they function, even if it makes them susceptible to things like adversarial attacks.\cite{DBLP:conf/nips/IlyasSTETM19} Deep neural networks are are not akin to scientists finding regularities and patterns and constructing theories capable of extrapolating to new scenarios. Any regularities that neural networks appear to have captured internally are solely due to the data that was fed to them, rather than a self-directed ``regularity extraction'' process. It is tempting to tell ``just-so'' stories on how a deep neural network is functioning, based on one of the explainability techniques mentioned previously. These stories can mislead from what they are actually doing - which is fitting a highly flexible function to do interpolation between nearby points. Specific details of the architecture are not important - performance is largely a function of depth and how densely and broadly the data samples the real-world distribution of inputs. 

%-------------------------------------------------------------- 
\section{Challenges in building trustworthy self-explaining AI}
\begin{figure*}[ht]
\includegraphics[width=\textwidth]{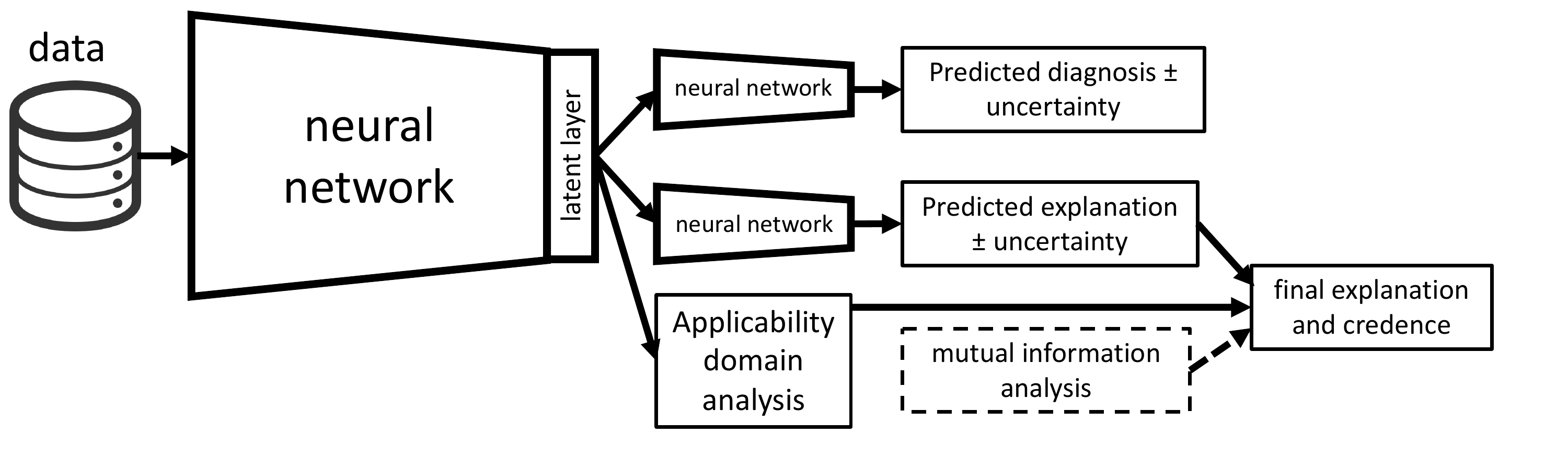}
\caption{Sketch of a simple self-explaining AI system.} \label{fig:diagram}
\end{figure*}

In his landmark 2014 book \textit{Superintelligence: Paths, Dangers, Strategies}, Nick Bostrom notes that highly advanced AIs may be incentivized to deceive their creators until a point where they exhibit a ``treacherous turn'' against them~\cite{Bostrom2014}. In the case of superintelligent or otherwise highly advanced AI, the possibility of deception appears to be a highly non-trivial concern. Here however, we suggest some methods by which we can trust the explanations given by present day deep neural networks, such as typical convolutional neural networks or transformer language models. Whether these methods will still have utility when it comes to future AI \& AGI systems is an open question. 

To show how we might create trust, we focus on an explicit and relatively simple example. Shen et al.~\cite{Shen2019} and later LaLonde et al.~\cite{LaLondearxiv} have both proposed deep neural networks for lung nodule classification which offer ``explanations''. Both authors make use of a dataset where clinicians have labeled lung nodules not only by severity (cancerous vs. non-cancerous) but also quantified them (on a scale of 1-5) in terms of five visual attributes which are deemed relevant for diagnosis (subtlety, sphericity, margin, lobulation, spiculation, and texture). While the details of the proposed networks vary, both output predictions for severity and scores for each of the visual attributes. Both authors claim that the visual attribute predictions ``explain'' the diagnostic prediction, since the diagnostic branch and visual attribute prediction branch(es) are connected near the base of the network. However, no evidence is presented that the visual attribute prediction is in any way related to the diagnosis prediction. While it may seem intuitive that the two output branches must be related, this must be rigorously shown for trustworthiness to hold.\footnote{Non-intuitive behaviours have repeatably been demonstrated in deep neural networks, for instance it has been shown networks based on rectified linear units contain unexpectedly large ``linear regions'' with many unused units inside them~\cite{HaninNIPS}.} %Hanlin2019ICML
Additionally, even if the visual attributes were used, no weights (``relevances'') are provided for the importance of each attribute to the prediction, and there may be other attributes of equal or greater importance that are used but not among those outputted (this point is admitted and discussed by Shen et al.~\cite{Shen2019}).  

Therefore, we would like to determine the degree to which the attributes in the explanation branch are responsible for the prediction in the diagnosis branch. We focus on the layer where the diagnosis and explanation branch diverge and look at how the output of each branch relates to activations in that layer. There are many ways of quantifying the relatedness of two variables, the Pearson correlation being one of the simplest, but also one of the least useful in this context since it is only sensitive to linear relationships. A measure which is sensitive to non-linear relationships and which has nice theoretical interpretation is the mutual information. For two random variables $X$ and $Y$ it is defined as: 
%\begin{widetext}
\begin{equation}
    \begin{aligned}
        \mbox{MI}(X,Y) &\equiv \sum_{y \in Y} \sum_{x \in X} p(x,y) \log{ \left(\frac{p(x,y)}{p(x) p(y)} \right) } \\
                % &= \sum_{y \in Y}\sum_{x \in X} p(x,y) \log{( p(x,y)) } - \sum_{y \in Y}\sum_{x \in X} p(x,y) \log{( p(x)) } - \sum_{y \in Y}\sum_{x \in X} p(x,y) \log{( p(y)) } \\
                % & = \sum_{y \in Y}\sum_{x \in X} p(x,y) \log{( p(x,y)) } - \sum_{x \in X} p(x) \log{( p(x)) } - \sum_{y \in Y} p(y) \log{( p(y)) } \\
                 & = H(x,y) - H(x) - H(y)
    \end{aligned}
\end{equation}
%\end{widetext}
Where $H(x)$ is the Shannon entropy. One can also define a mutual information correlation coefficient:\cite{Linfoot1957:85}
\begin{equation}
    r^{\ff{MI}}(X,Y) = \sqrt{ 1 - e^{-2 \mbox{\ff{MI(X,Y)}}} }
\end{equation}
This coefficient has the nice property that it reduces to the Pearson correlation in the case that $P(x,y)$ is a Gaussian function with non-zero covariance. The chief difficulty of applying mutual information is that the underlying probability distributions $P(x,y)$, $P(x)$, and $P(y)$ all have to be estimated. Various techniques exist for doing this however, such as by using kernel density estimation with Parzen windows~\cite{Torkkola2003:1415}.\footnote{Note that this sort of approach should not be taken as quantifying ``information flow'' in the network. In fact, since the output of units is continuous, the amount of information which can flow through the network is infinite (for discussion and how to recover the concept of ``information flow'' in neural networks see~\cite{Chaudhuri2019informationflow}). What we propose to measure is the the mutual information over the data distribution used.}

Suppose the latent vector is denoted by $\bs{L}$ and has length $N$. Denote the diagnosis of the network as $D$ and the vector of attributes $\bs{A}$. Then for a particular attribute $A_j$ in our explanation word set we calculate the following to obtain a ``relatedness'' score between the two:
\begin{equation}
    R(A_j) = \sum\limits_i^N \mbox{MI}(L_i, D)\mbox{MI}(L_i, A_j) 
\end{equation}

An alternative (an perhaps complimentary) method is to train a ``post-hoc'' model to try to predict the diagnosis from the attributes (also shown in figure~\ref{fig:diagram}). While this cannot tell us much about mechanism of the main model (due to the Rashomon effect) we can learn a bit from it. Namely, if the post-hoc model is not as accurate as the diagnosis branch of the main model, then we know the main model is using additional features. 

A final alternative is to have a completely separate algorithm automatically derive the explanation. In a similar spirit to the ``auto-ML'' movement, this could be called ``auto-explanation''. Ideally the system would have access to the full model, since the Rashomon effect noted earlier suggests many models with very different mechanism can have the same input-output mappings. However, it's possible this may not be the case due to intellectual property reasons.  Developing mathematically rigorous techniques for ``shining lights'' into ``black boxes'' was a popular topic in early cybernetics research~\cite{ashbycybernetics}. With some assumptions as to the architecture of the model recently it has been shown that weights can be inferred for ReLU networks through careful analysis of input-output relations~\cite{rolnick2019identifying}. 
 
%-------------------------------------------------------------- 
\section{Ensuring robustness through applicability domain and uncertainty analysis}
The concept of an ``applicability domain'', or the domain where a model makes good predictions, is well studied in the area of molecular modeling known as quantitative structure property relationships (QSPR), and practitioners in that field have developed a number of techniques which are ready for export (for a review, see~\cite{Sahigara2012} or~\cite{Netzeva2005}). It is remarkable that quantifying the applicability domain of models hasn't become more widespread in other areas where machine learning, given concerns about robustness and adversarial attacks. An analysis of applicability domain analysis methods for deep learning and in particular computer vision is outside the scope of this paper and will be the subject of a future work. However, as an illustration, one way of delineating the applicability domain is to calculate the convex hull of the input vectors for all training data points (if the input is very high dimensional, dimensionality reduction should be applied first). If the input/latent vector of a test data point falls outside the convex hull, then the model should send an alert saying that the test point falls outside the model's applicability domain. Applicability domain analysis can be framed as a simple form of AI self-awareness, which is thought by some to be an important component for AI safety in advanced AIs~\cite{Aliman2018AGIconf}.

Finally, models should contain measures of uncertainty for both their decisions and their explanations. Ideally, this should be done in a Bayesian way using a Bayesian neural network~\cite{Garin2017uncertainty}. With the continued progress of Moore's law, training Bayesian CNNs~\cite{McClure2019} is now becoming feasible and in our view this is a worthwhile use of additional CPU/GPU cycles. There are also approximate methods - for instance it has been shown that random dropout during inference can be used to estimate uncertainties at little extra computational cost~\cite{Gal2016DropoutUncertainty}. Just as including experimental error bars is standard in all of science, and just as we wouldn't trust a doctor who could not also give a confidence level in his diagnosis, uncertainty quantification should be standard practice in AI research. 

%-------------------------------------------------------------- 
\section{Conclusion}
We argued that deep neural networks trained on complex real world data are very difficult to interpret due to their power arising from brute-force interpolation over big data rather than through the extraction of high level generalizable rules. Motivated by this and by the need for trust in AI systems we introduced the concept of self-explaining AI and described how a simple self-explaining AI would function for diagnosing medical images. To build trust, we showed how a mutual information metric can be used to verify that the explanation given is related to the diagnostic output. Crucially, in addition to an explanation, self-explaining AI outputs confidence levels for both the decision and explanation, further aiding our ability to gauge the trustworthiness of any given diagnosis or decision. Finally, an applicability domain analysis should be done for AI systems where robustness and trust are important, so that systems can alert their user if they are asked work outside their domain of applicability. 
\\\\
%-------------------------------------------------------------- 
\textbf{Funding \& disclaimer}\\
No funding sources were used in the creation of this work. The author (Dr.\ Daniel C.\ Elton) wrote this article in his personal capacity. The opinions expressed in this article are the author's own and do not reflect the view of the National Institutes of Health, the Department of Health and Human Services, or the United States government.

% Acknowledgements should only appear in the accepted version.
\section*{Acknowledgements}
The author appreciates the helpful feedback from Robert Kirk, Kyle Vedder, Jacob Reinhold, Dr.\ W. James Murdoch, and Dr.\ Ulas Bagci.
% In the unusual situation where you want a paper to appear in the
% references without citing it in the main text, use \nocite
%\nocite{langley00}
\bibliography{bibliography}
\bibliographystyle{icml2020}
\end{document}